\documentclass{article}
\usepackage{ijcai25}
\usepackage[table,dvipsnames]{xcolor} 
\usepackage{booktabs} 
\usepackage{times}
\usepackage{soul}
\usepackage{url}
\usepackage[hidelinks]{hyperref}
\usepackage{tcolorbox}

\usepackage[utf8]{inputenc}
\usepackage[small]{caption}
\usepackage{graphicx}
\usepackage{amsmath}
\usepackage{amsthm}
\usepackage{algorithm}
\usepackage{algorithmic}
\usepackage{bbding}
\usepackage[switch]{lineno}
\usepackage{multirow}
\usepackage{float}
\usepackage{afterpage}
\usepackage{footmisc}  

\definecolor{LightBlue}{rgb}{0.8,0.9,1.0}
\definecolor{LightYellow}{rgb}{1.0,1.0,0.8}
\definecolor{LightGray}{rgb}{0.9,0.9,0.9}
\definecolor{LightOrange}{rgb}{1.0,0.9,0.8}

\definecolor{highlightgreen}{HTML}{D9F2D9}

\definecolor{cvprblue}{rgb}{0.21,0.49,0.74}
\definecolor{tractorred}{rgb}{0.99, 0.05, 0.21}
\definecolor{tealgreen}{rgb}{0.0, 0.51, 0.5}

\title{Localizing Before Answering: A Benchmark for Grounded Medical Visual Question Answering}

\author{
Dung Nguyen$^1$ \and 
Minh Khoi Ho$^1$ \and 
Huy Ta$^2$ \and 
Thanh Tam Nguyen$^3$ \and 
Qi Chen$^2$ \and 
Kumar Rav$^4$ \and 
Quy Duong Dang$^2$ \and 
Satwik Ramchandre$^2$ \and 
Son Lam Phung$^6$ \and 
Zhibin Liao$^2$ \and 
Minh-Son To$^5$ \and 
Johan Verjans$^2$ \and 
Phi Le Nguyen$^1$ \and 
Vu Minh Hieu Phan$^2$\\
\affiliations
$^1$ Hanoi University of Science and Technology \\
$^2$ Australian Institute for Machine Learning, The University of Adelaide\\
$^3$ Griffith University \\ $^4$ SA Health \\
$^5$ College of Medicine and Public Health, Flinders University\\
$^6$ University of Wollongong\\
\emails
}

\begin{document}
\maketitle
\vspace{-5mm}
\begin{abstract}
Medical Large Multi-modal Models (LMMs) have demonstrated remarkable capabilities in medical data interpretation. However, these models frequently generate hallucinations contradicting source evidence, particularly due to inadequate localization reasoning. This work reveals a critical limitation in current medical LMMs: instead of analyzing relevant pathological regions, they often rely on linguistic patterns or attend to irrelevant image areas when responding to disease-related queries.
To address this, we introduce HEAL-MedVQA (Hallucination Evaluation via Localization MedVQA), a comprehensive benchmark designed to evaluate LMMs' localization abilities and hallucination robustness. HEAL-MedVQA features (i) two innovative evaluation protocols to assess visual and textual shortcut learning, and (ii) a dataset of 67K VQA pairs, with doctor-annotated anatomical segmentation masks for pathological regions. To improve visual reasoning, we propose the Localize-before-Answer (LobA) framework, which trains LMMs to localize target regions of interest and self-prompt to emphasize segmented pathological areas, generating grounded and reliable answers. Experimental results demonstrate that our approach significantly outperforms state-of-the-art biomedical LMMs on the challenging HEAL-MedVQA benchmark, advancing robustness in medical VQA. 
Code is available \fcolorbox{blue}{white}{\href{https://github.com/tuandung2812alt3/Localize-before-Answering/}{here}}.
\end{abstract}




\section{Introduction}
\label{sec:intro}

Large Multimodal Models (LMMs), or Multi-modal Large Language Models (M-LLMs)~\cite{alayrac2022flamingo,bai2023qwenvl,liu2024visual} such as GPT-4V~\cite{gpt4} achieve superb performance in understanding data from multiple modalities, such as vision and language, and generating human-like texts. Leveraging the strong capabilities of those foundational models, recent works~\cite{li2024llava,chen2024chexagent,zhang2023biomedgpt,chen2023medblip,moor2023med,wu2023towards,wu2023medklip,yan2022clinical} have developed M-LLMs for medical imaging applications such as medical Visual Question Answering (VQA). Recently, many multimodal foundation models (LLaVA-Med \cite{li2024llava},  CheXagent \cite{chen2024chexagent}, GPT-4 \cite{gpt4}, Gemini \cite{gemini}) have emerged and demonstrated impressive reasoning and comprehension capabilities on biomedical queries and related domains. 
These models can assist clinical professionals in interpreting and gaining insights on medical images, helping the diagnosis and treatment processes be more efficient. 


\begin{figure*}[h!]
\centering
\includegraphics[width=0.95\linewidth]{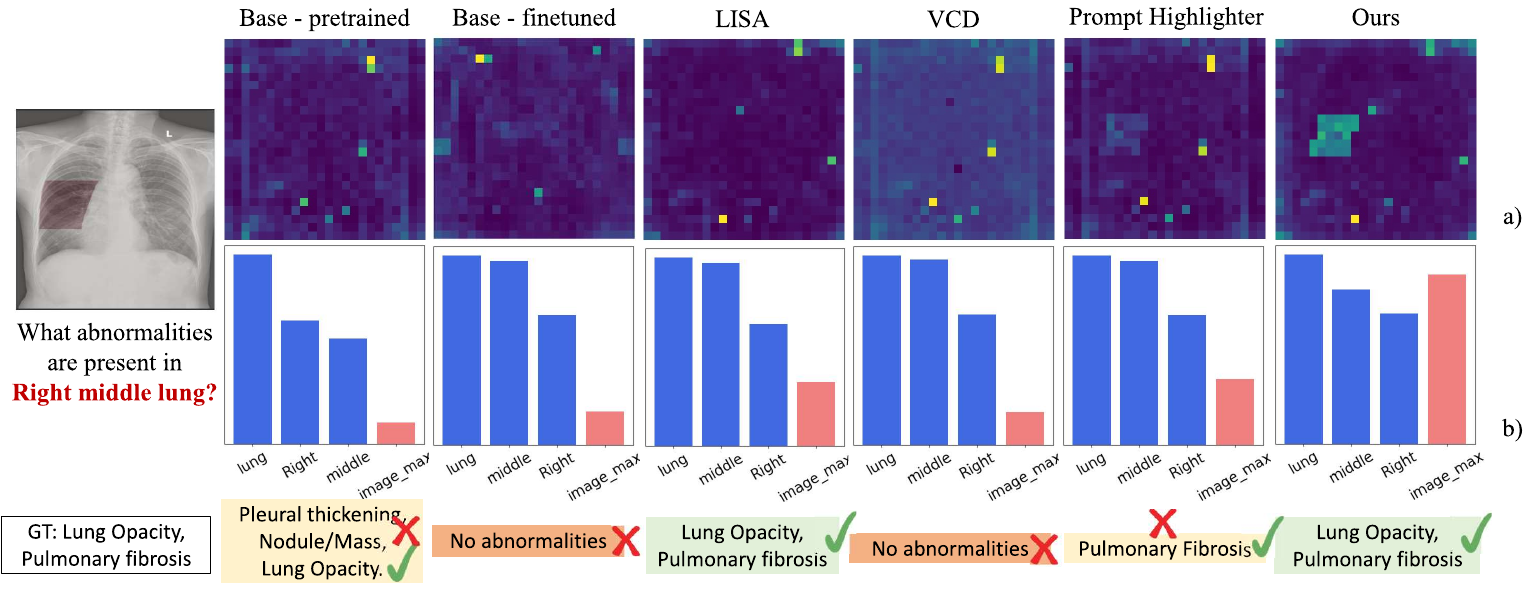}
\caption{An example of Med-VQA from our dataset. We follow LVLM-Interpret \protect\cite{stan2024lvlm} to visualize a) the attention map of the image with respect to the model response, and b) the relevancy score between the answer and input words. LLaVA-Med \protect\cite{li2024llava}, as reference model, fails to attend to the region of interest, and its answer has a low relevancy score to the image. Several hallucination reduction methods can generate responses more relevant to the image, but may not localize the target region.}
\label{fig:intro}
\end{figure*}

Due to the sensitive nature of the biomedical domain, the accuracy and trustworthiness of these foundational models are important. Though impressive, Large Language Models (LLMs) and multimodal LLMs are prone to hallucinating contents. Specifically, the models produce false answers, which are not grounded in the visual evidence.   
Given a question. \textit{``Does the patient have pneumonia in his right lung?"}, many LLMs are prone to answer \textit{``Yes"} without reasoning the image content. This behavior arises because the observation of \textit{pneumonia} frequently co-occurs with parts of the lung, leading to spurious correlations instead of rigorous reasoning.
Fig.~\ref{fig:intro} illustrates the answer of some open-source LMMs, and the cross-attention map of the models during generation. 
These models tend to hallucinate and attend to irrelevant image regions when answering these questions.


This study shows two types of shortcut learning that current multimodal LLMs suffer from: textual and visual shortcut learning. \textit{First}, the models attend to non-important text tokens with higher relevancy score than any image tokens. As shown in Fig.~\ref{fig:intro}b, the model produces higher relevancy scores on the token \textit{lung} than on any image token. \textit{Second}, on the image, the model attends to non-queried image regions (as shown in Fig.~\ref{fig:intro}a). This shows the model learns shortcuts on irrelevant text and image tokens, instead of basing on the visual evidence when answering the questions.

To address these concerns, we introduce a new benchmark, dubbed HEAL-MedVQA (Hallucination EvAluation via Localization in Medical VQA),
to dissect the LLM's ability to localize at visual evidence when answering. Specifically, this paper proposes two evaluation protocols, called Textual Perturbation Test (TPT) and Visual Perturbation Test (VPT), which probe LLMs' sensitivity towards textual and visual shortcuts. To support this evaluation, we curate 
adversarial VQAs and procure pixel-level annotations of affected anatomy regions from three board-certified radiologists from two large-scale public datasets, MIMIC-CXR \cite{johnson2019mimic} and VinDr-CXR \cite{nguyen2020vindrcxr}. 
In total, we construct over 60,000 question-answer pairs probing LLM's textual and visual shortcut learning. Our curated medical benchmark provides a structured codebase, facilitating comprehensive comparisons of advanced LLMs. 

To this end, we propose 
Localize-before-Answer (LobA), a framework that forces LMM to generate visual-evidence tokens, which is then input to the segmentation decoder to localize the area of interest before answering.
Based on the localization map, LobA self-prompts to enhance the model's attention, forcing it to leverage visual evidence relevant to the question before answering. 
In summary, our main contributions are as follows.
\begin{itemize}
    \item We introduce HEAL-MedVQA  (Hallucination Evaluation via Localization in Medical VQA), a benchmark that consists of 67K QA pairs, and doctor-annotated segmentation masks with two new evaluation protocols, assessing LLMs' robustness to textual and visual shortcut learning. Our benchmark includes comprehensive comparisons of 8 state-of-the-art LLMs and hallucination techniques, thus standardizing hallucination and accuracy evaluations of future multi-modal LLMs.
    \item We propose the Localize-before-Answer (LobA) framework, which forces the model to localize the affected anatomical regions and self-prompt to enhance attention on the target regions before answering.
    \item Our proposed LobA framework significantly surpasses recent LLMs by up to 5.44\%, while showing the highest robustness to textual and visual shortcut learning.
\end{itemize}

\section{Related Works}
\label{sec:related}

\textbf{Medical Visual Question Answering.} 
Several benchmarks in Med-VQA propose specialized datasets~\cite{Hasan2018OverviewOI,Abacha2019VQAMedOO,Abacha2020OverviewOT,Abacha2021OverviewOT}. 
VQA-RAD \cite{lau2018vqa} offers both open-ended and closed-ended questions. SLAKE~\cite{liu2021slake} develops a bilingual dataset with a knowledge graph. 
Overall, existing benchmarks use accuracy as the main metric, which overlooks the hallucination evaluation of  multi-modal LLMs.

A majority of Med-VQA models are finetuned from other VLMs, namely LLaVA-Med \cite{li2024llava} from LLaVA \cite{liu2024visual}, CheXagent \cite{chen2024chexagent} from BLIP-2 \cite{li2023blip}, or \cite{zhang2023biomedgpt,chen2023medblip,moor2023med,wu2023towards,wu2023medklip,yan2022clinical,huy2025seeing}. 
Recent approaches, which localize relevant regions in Med-VQA, have gained attention in recent times \cite{tascon2023localized,tascon2024targeted,zhang2024dual}. 
However, during inference time, these VQA models require human-annotated masks to locate regions of interest. In contrast, our method explicitly learns to localize during training, bypassing the need to acquire human annotations when inferring.  

\noindent \textbf{Hallucination in VQA.} 
To overcome hallucination, there are two main approaches. \textit{First}, data-centric methods 
focus on augmenting existing VQA training datasets and propose new benchmarks for evaluation. In \cite{li2023evaluating} and \cite{liu2023mitigating}, 
they show that frequently appearing objects tend to guide models' answers during inference time more than absent or less frequent ones. As such, POPE~\cite{li2023evaluating} adds adversarial questions asking for the presence of objects that are absent from the image. 
LRV-Instruction \cite{liu2023mitigating} introduces negative instructions, which are irrelevant to the visual content. 

\textit{Second}, 
model-centric methods focus on modifying the model architecture to support the visual component. Intern-VL \cite{chen2024internvl}, and LLaVA-1.5-HD \cite{liu2024improved} scale up the vision foundation model to allow better visual understanding. Others, such as \cite{he2024incorporating,jain2024vcoder} append several levels of information (segmentation mask, object list, task-based knowledge) to enrich the visual context. 
Lastly, several methods propose inference-based methods, which augment the inference of the pre-trained LLMs. 
VCD~\cite{leng2024mitigating} and M3ID \cite{favero2024multi} aim to ground inference by negating a contrastive version of the input where the image is distorted or dropped. OPERA \cite{huang2024opera} discourages overly confident but erroneous information accumulated in self-attention layers by applying a penalty. Overall, existing hallucination techniques are applied on the image level, while neglecting region-level details. 
 In contrast, we propose to explicitly reason on the pixel level before answering.

\section{Hallucination Evaluation Benchmark}
This section introduces HEAL-MedVQA  (Hallucination Evaluation via Localization in Medical VQA) benchmark consists of 67K QA pairs, and doctor-annotated segmentation masks with two new evaluation protocols.
\subsection{New Evaluation Protocol — Sensitivity Test}
\label{sec:protocol}

\begin{figure}[!h]
    \centering
    \includegraphics[width=0.92\linewidth]{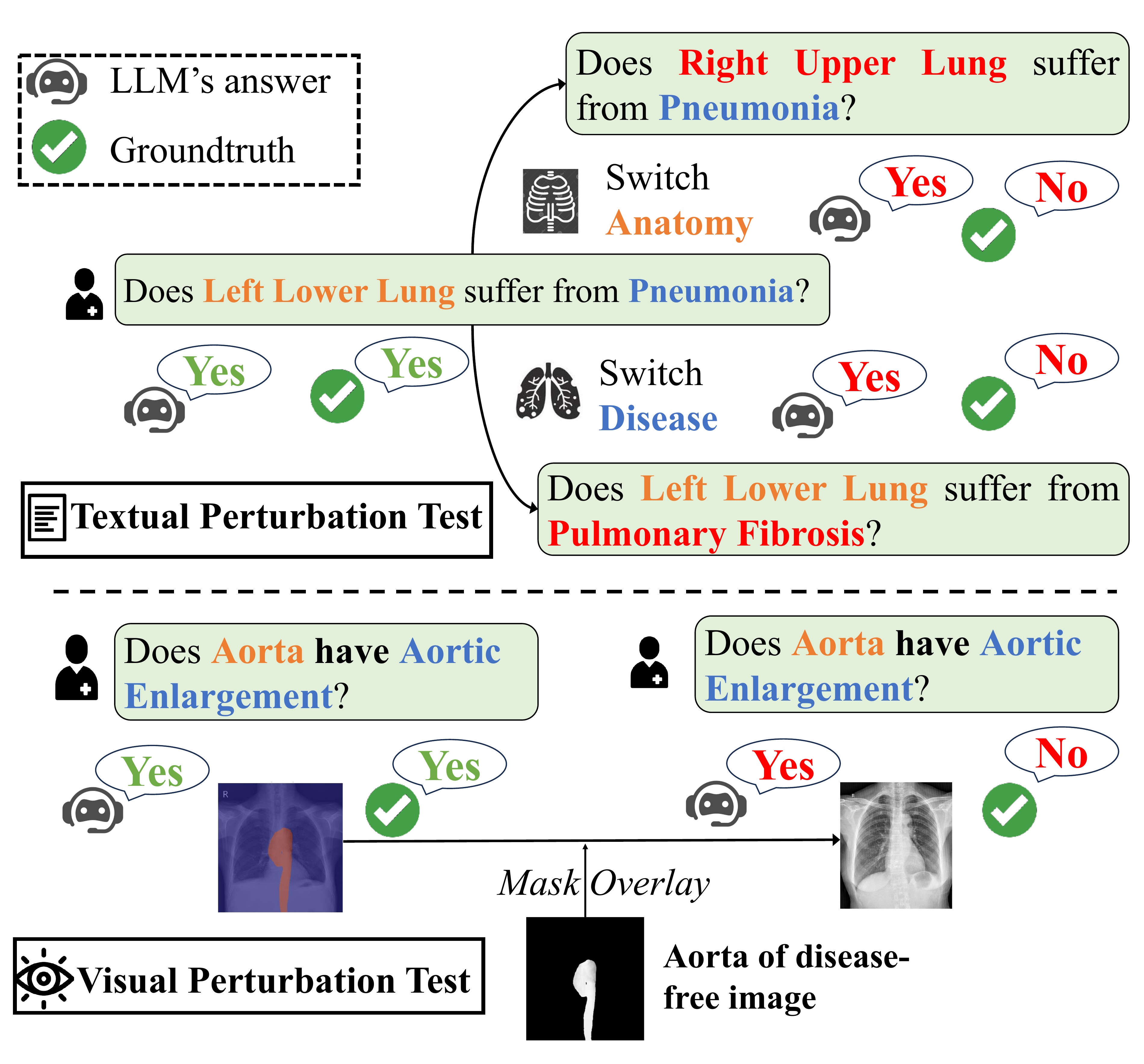} 
    \caption{Textual and Visual Perturbation Tests.
TPT: Replace the anatomy or disease term in the question.
VPT: Swap the region of interest with the same region from an image of a different class.
We report the percentage of “Yes” → “No” flips to assess model sensitivity to shortcuts.}
    \label{fig:sensitivity_analsyis}
\end{figure}

Conventional accuracy metrics~\cite{Abacha2019VQAMedOO,liu2021slake,lau2018vqa}  overlook bias factors influencing the model's responses. 
This paper assesses two key phenomena in medical LMMs:
\begin{itemize}
    \item \textbf{Language Shortcut:} Despite generating correct answers, the model 
    relies on language shortcuts learnt in the training dataset, rather than the actual visual content. 
    For example, given a question \textit{``Does the right lower lung have pneumonia?"}, models can answer \textit{``Yes"} due to the co-occurrence of  \textit{lung} and \textit{pneumonia} in the training set.
    \item 
    \textbf{Vision Shortcut:} 
    The model can concentrate on non-relevant regions (as in Fig.~\ref{fig:intro}a). This behavior results in the unreliability of LMMs, even when providing correct answers, as they ignore visual evidence.
\end{itemize}
To this end, we propose two new evaluation protocols, 
called Textual Perturbation Test (TPT) and Visual Perturbation Test (VPT), as shown in Fig.~\ref{fig:sensitivity_analsyis}.

\noindent \textbf{Textual Perturbation Test.} 
To assess how sensitive a model is to language biases, we propose to swap the key entities, \textit{i.e.}, anatomy or disease, and evaluate the model's sensitivity to the perturbed questions. The evaluation protocol is as follows. 
\begin{itemize}
    \item From the test set of binary questions - “Does \textit{anatomy} have \textit{disease}?”, we collect True Positive samples having correct “Yes” predictions. 
    \item We randomly replace either the anatomy or disease with another valid co-occurring term. 
    For example, in Fig. \ref{fig:sensitivity_analsyis} (Above), the question \textit{``Does left lower lung suffer from pneumonia?"} is replaced by either switching the anatomy to \textit{right upper lung}, or to another disease.
    \item The percentage of “Yes” → “No” flips defines the \textit{Textual Perturbation Test} (TPT) score. 
\end{itemize}



\noindent \textbf{Visual Perturbation Test.}
We propose to assess whether the model grounds the answer on the anatomical region of interest. 
Specifically, when querying the presence of a disease in a particular area, \textit{e.g.}, the left lower lung, we take a segmented region of the left lower lung from another image that does not share the same diagnosis as the current one. This segmented region is resized and blended onto the original image. By doing so, we expect the model to focus its attention on the replaced region of interest to produce the correct diagnosis, altering the response from ``Yes" to ``No". We define the visual perturbation test (VPT) score as the \textit{percentage of answers that change} from ``Yes" to ``No" under this protocol.



\subsection{Dataset Curation}

To assess the LLM's robustness against both textual and visual shortcuts, we introduce a new benchmark called HEAL-MedVQA (Hallucination Evaluation via Localization in Medical VQA).
Built upon VinDr-CXR \cite{nguyen2020vindrcxr} and MIMIC-CXR \cite{johnson2019mimic}, we construct HEAL-MedVQA by mapping spatial relationships between diseases and anatomical locations, and probing the model's robustness towards disease and location perturbation on both texts and images. Fig.~\ref{fig:data-pipeline} illustrates our overall pipeline of data construction. We extract the disease bounding boxes and anatomical masks from the image. 
Based on the overlap between disease and anatomy localization, we determine their spatial relationships. Then, several QA templates are used to generate VQA samples. This process can be broken down into four main steps:

\begin{figure}[!h]
    \centering
    \includegraphics[width=\linewidth]{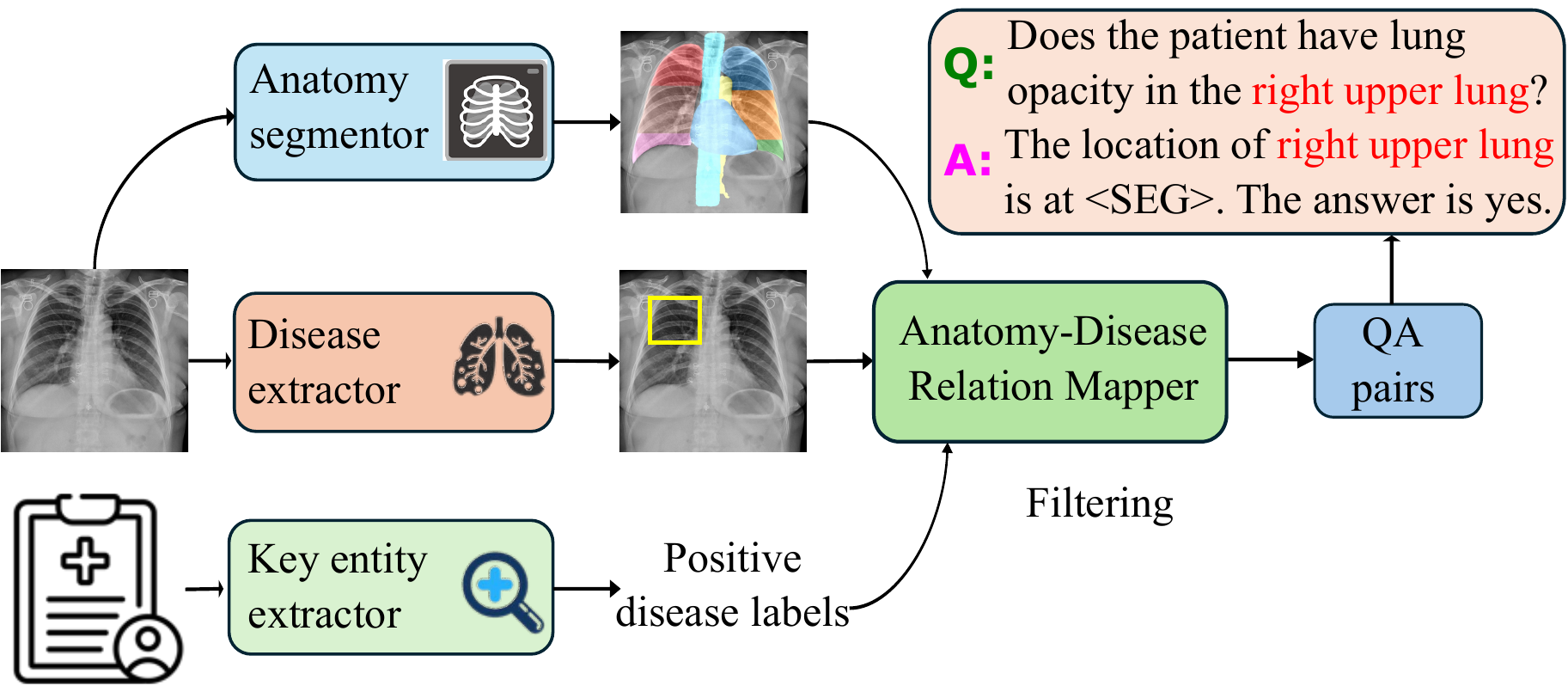}
    \caption{Our data processing pipeline.}
    \label{fig:data-pipeline}
\end{figure}


\textit{Anatomy segmentation.} We define 8 main anatomies, e.g. \textit{Right lower lung, Heart, etc.}, where chest diseases commonly occur. On the training set, we obtain the segmentation mask using a model pre-trained on PAXRay++ dataset~\cite{seibold2022detailed}. On the test set, we consult three board-certified radiologists to annotate the segmentation. By providing doctor-annotated pixel-level segmentation, this research offers a new benchmark to evaluate visual reasoning and robustness to hallucination of current M-LLMs.

\textit{Disease extraction.} For the disease bounding boxes, the VinDr-CXR dataset provides bounding box annotations for 22 medical findings. We train a standard detection model, YOLOv5 \cite{glennjocher20204154370}, on the dataset and obtain the bounding boxes on MIMIC-CXR. To reduce false positives, we extract the disease labels from the associated medical reports and filter out absent diseases.

\textit{Anatomy-disease relation mapping.} To get the anatomy-disease relation, we measure the overlap between the disease bounding box and the anatomical segmentation mask. Since the disease area is often small compared to the anatomy's, we compute the intersection over the disease area, instead of a union of both.  
If this IoU$_\text{dis}$ score between a disease-anatomy pair is over $\delta=0.5$, then that disease occurs on that anatomy. 

\textit{Question Generation.} After obtaining a list of disease–anatomy mappings, we randomly generate 2–5 QA pairs per image, focusing on two main formats: close-ended and open-ended, as shown in Fig.~\ref{fig:data-pipeline}. Close-ended questions are binary: ``Does \textit{anatomy} have \textit{disease}?", ``Is there \textit{disease} present in \textit{anatomy}?". Each is either a Positive question, which queries a disease present in the anatomy, or a Hallucinated question, which queries a disease that is \textbf{not} present.
Open-ended questions have two types: normal and abnormality-query questions, which respectively probe regions without and with diseases. 

\begin{figure}[h]
    \centering
    \includegraphics[width=1.0\linewidth]{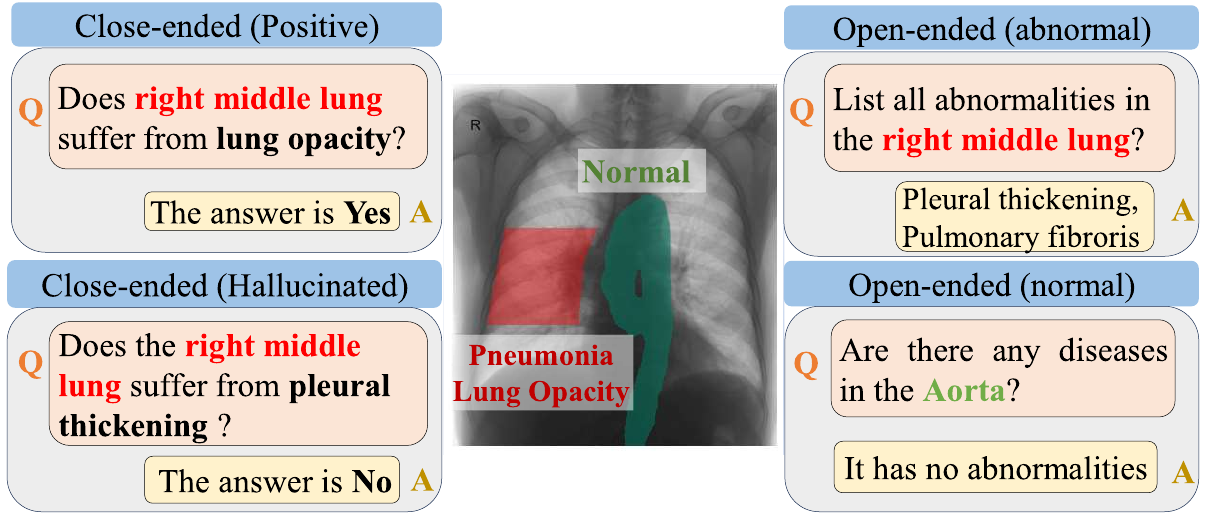}
    \caption{Four question types in our benchmark.}
    \label{fig:data-pipeline}
\end{figure}




In total, our HEAL-MedVQA dataset comprises 45,331 closed-ended questions and 22,573 open-ended ones. 
We compare our benchmark with those previously reported in Table \ref{tab:dataset_characteristics}. To the best of our knowledge, we are the first benchmark providing a large-scale dataset with comprehensive hallucination metrics and doctor-annotated masks to evaluate the hallucination of emerging LLMs.

\begin{table*}[h]
\hspace{2mm}\footnotesize
\begin{tabular}{p{4.2cm}p{0.2cm}p{0.3cm}p{0.3cm}p{0.3cm}p{0.4cm}p{0.4cm}p{0.4cm}p{0.4cm}} 
\hline\hline
\textbf{Datasets} & \multicolumn{1}{c}{\textbf{\# Images}} & \multicolumn{1}{c}{\textbf{{\begin{tabular}[c]{@{}c@{}}\# QA\\pairs\end{tabular}}}} & \multicolumn{1}{c}{\textbf{\begin{tabular}[c]{@{}c@{}}Adversarial\\ questions\end{tabular}}} & \multicolumn{1}{c}{\textbf{Closed-ended}} & \multicolumn{1}{c}{\textbf{Open-ended}} & \multicolumn{1}{c}{\textbf{\begin{tabular}[c]{@{}c@{}}Visual \\ grounding\end{tabular}}} & \multicolumn{1}{c}{\textbf{Annotation}} \\ \hline
VQA-Rad \cite{lau2018vqa} & \multicolumn{1}{c}{315} & \multicolumn{1}{c}{3.5K} & \multicolumn{1}{c}{\textcolor{red}{\XSolidBrush}} & \multicolumn{1}{c}{\textcolor{Green}{\CheckmarkBold}} & \multicolumn{1}{c}{\textcolor{Green}{\CheckmarkBold}} & \multicolumn{1}{c}{\textcolor{red}{\XSolidBrush}} & \multicolumn{1}{c}{\textcolor{red}{\XSolidBrush}}  \\
SLAKE \cite{liu2021slake} & \multicolumn{1}{c}{642} & \multicolumn{1}{c}{14K} & \multicolumn{1}{c}{\textcolor{red}{\XSolidBrush}} & \multicolumn{1}{c}{\textcolor{Green}{\CheckmarkBold}} & \multicolumn{1}{c}{\textcolor{Green}{\CheckmarkBold}} & \multicolumn{1}{c}{\textcolor{red}{\XSolidBrush}$^{\dagger}$} & \multicolumn{1}{c}{Mask/Bbox} \\
VQA-Med \cite{Abacha2021OverviewOT} & \multicolumn{1}{c}{5.5K} & \multicolumn{1}{c}{5.5K} & \multicolumn{1}{c}{\textcolor{red}{\XSolidBrush}} & \multicolumn{1}{c}{\textcolor{red}{\XSolidBrush}} & \multicolumn{1}{c}{\textcolor{Green}{\CheckmarkBold}} & \multicolumn{1}{c}{\textcolor{red}{\XSolidBrush}} & \multicolumn{1}{c}{\textcolor{red}{\XSolidBrush}} \\
PMC-VQA \cite{zhang2023pmcvqa} & \multicolumn{1}{c}{149K} & \multicolumn{1}{c}{227K} & \multicolumn{1}{c}{\textcolor{red}{\XSolidBrush}} & \multicolumn{1}{c}{\textcolor{Green}{\CheckmarkBold}} & \multicolumn{1}{c}{\textcolor{Green}{\CheckmarkBold}} & \multicolumn{1}{c}{\textcolor{red}{\XSolidBrush}} & \multicolumn{1}{c}{\textcolor{red}{\XSolidBrush}} \\
OmniMedVQA \cite{hu2024omnimedvqa} & \multicolumn{1}{c}{118k} & \multicolumn{1}{c}{128K} & \multicolumn{1}{c}{\textcolor{red}{\XSolidBrush}} & \multicolumn{1}{c}{\textcolor{Green}{\CheckmarkBold}} & \multicolumn{1}{c}{\textcolor{Green}{\CheckmarkBold}} & \multicolumn{1}{c}{\textcolor{red}{\XSolidBrush}} & \multicolumn{1}{c}{\textcolor{red}{\XSolidBrush}} \\
CARES \cite{xia2024cares} & \multicolumn{1}{c}{18K} & \multicolumn{1}{c}{41K} & \multicolumn{1}{c}{\textcolor{red}{\XSolidBrush}} & \multicolumn{1}{c}{\textcolor{Green}{\CheckmarkBold}} & \multicolumn{1}{c}{\textcolor{Green}{\CheckmarkBold}} & \multicolumn{1}{c}{\textcolor{red}{\XSolidBrush}} & \multicolumn{1}{c}{\textcolor{red}{\XSolidBrush}} \\ \hline
Halt-MedVQA \cite{wu2024hallucination} & \multicolumn{1}{c}{1736} & \multicolumn{1}{c}{2359} & \multicolumn{1}{c}{\textcolor{Green}{\CheckmarkBold}} & \multicolumn{1}{c}{\textcolor{Green}{\CheckmarkBold}} & \multicolumn{1}{c}{\textcolor{red}{\XSolidBrush}} & \multicolumn{1}{c}{\textcolor{red}{\XSolidBrush}} & \multicolumn{1}{c}{\textcolor{red}{\XSolidBrush}} \\
ProbMed \cite{yan2024worse} & \multicolumn{1}{c}{6.3K} & \multicolumn{1}{c}{57K} & \multicolumn{1}{c}{\textcolor{Green}{\CheckmarkBold}} & \multicolumn{1}{c}{\textcolor{Green}{\CheckmarkBold}} & \multicolumn{1}{c}{\textcolor{red}{\XSolidBrush}} & \multicolumn{1}{c}{\textcolor{red}{\XSolidBrush}} & \multicolumn{1}{c}{\textcolor{red}{\XSolidBrush}} \\
\rowcolor{green!20} \textbf{HEAL-MedVQA (Ours)} & \multicolumn{1}{c}{34K} & \multicolumn{1}{c}{67K} & \multicolumn{1}{c}{\textcolor{Green}{\CheckmarkBold}} & \multicolumn{1}{c}{\textcolor{Green}{\CheckmarkBold}} & \multicolumn{1}{c}{\textcolor{Green}{\CheckmarkBold}} & \multicolumn{1}{c}{\textcolor{Green}{\CheckmarkBold}} & \multicolumn{1}{c}{Mask} \\ \hline\hline
\end{tabular}
\caption{Comparison of Med-VQA datasets. $^{\dagger}$ indicates visual annotations for all anatomic regions in the image instead of the one questioned.}
\label{tab:dataset_characteristics}
\end{table*}

\section{Proposed Method}

\label{sec:method}

\begin{figure*}[h]
    \centering
   \includegraphics[width=1.0\textwidth]{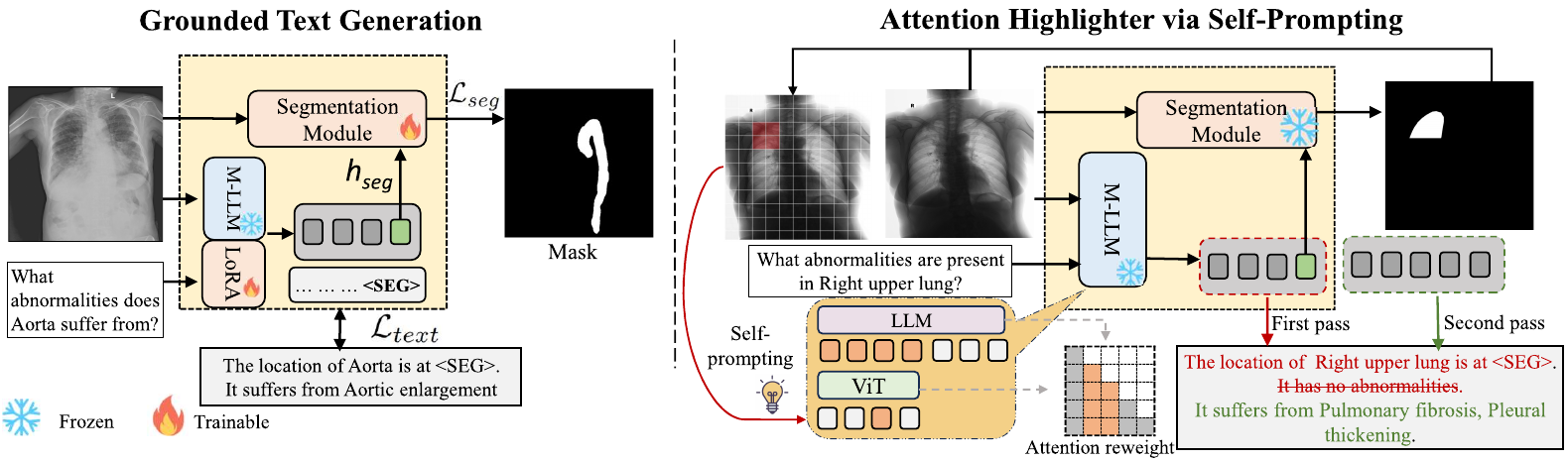}
   \caption{The proposed LoBA frameworks consists of two phases. \textbf{Grounded text generation}: during training, our model learns to localize and segment a region of interest mentioned in the question. \textbf{Self-prompting}: during inferencing, the model attention is calibrated to attend to the segmented region. With the attention, the model refines and generates visually grounded answers that are more robust to hallucinations.}
   \label{fig:arch}
\end{figure*}
This section presents our proposed simple-yet-effective framework, \textbf{Lo}calize-\textbf{b}efore-\textbf{A}nswer (\textbf{LobA}) to (i) localizes the queried regions, and then (ii) reweigh the attention to ground on correct visual evidence before answering.  
An overview of the framework is shown in Figure \ref{fig:arch}.

\subsection{Grounded Text Generation}
Given an image $x_{img}$ and a question about it $x_{text}$. 
We train LLMs to generate an answer $\hat{y}_{text}$, while grounding on the queried region, and producing a segmentation mask $\hat{M}$. 
The mask $\hat{M}$ corresponds to both $x_{text}$ and $\hat{y}_{text}$, explaining the response of the model.

We use a \textit{\textless SEG\textgreater} token, denoting the token that will be used in later steps to obtain the segmentation of the relevant anatomies. 
Using this token, we instruct LLMs to localize on visual evidence before answering. As such, we format the response as   
\textit{``The location of \{relevant areas\} is at \textless SEG\textgreater"}. For example, if the question is \textit{``Are there any abnormalities at the right lower lung?"}, the output of the LMM will be \textit{``The location of the right lower lung is at \textless SEG\textgreater"}. Note that, in our format, the model has to produce the visual evidence first, before giving the final answers. The LLM is trained to generate the above response:
\begin{equation}
    \hat{y}_{\text{text}} = LLM(x_{\text{text}},x_{\text{img}}).
\end{equation}
We use the projected hidden state $h_{\text{seg}}$ from the \textit{\textless SEG\textgreater} token as the query to the segmentation model. This can be formulated as $\hat{M} =  Seg(x_{\text{img}},h_{\text{seg}})$.
Our LLM is trained via the text objectives. Similar to other LLMs \cite{touvron2023llama,gpt4}, we formulate the text generation loss as the cross-entropy loss between the generated text and the ground truth:
\begin{equation}
    \mathcal{L}_{text} = \text{CE}(\hat{y}_{\text{text}},y_{\text{text}}).
\end{equation}
We combine BCE loss and DICE loss to train the segmentation module:
\begin{equation}
    \mathcal{L}_{seg} = \lambda_{\text{BCE}} \text{BCE}(\hat{M},M) + \lambda_{\text{DICE}} \text{DICE}(\hat{M},M).
\end{equation}
Finally, the overall objective can be seen as the weighted sum of the two losses:
\begin{equation}
    \mathcal{L} = \lambda_{\text{text}} \mathcal{L}_{\text{text}} + \lambda_{\text{seg}} \mathcal{L}_{\text{seg}}.
\end{equation}

\afterpage{%
\begin{table*}[!htbp]
\centering
\renewcommand{\arraystretch}{1.0} 
\setlength{\tabcolsep}{4pt} 
\resizebox{\textwidth}{!}{%
\begin{tabular}{l|ccc|ccc|ccc|ccc}
\toprule
\multirow{3}{*}{\textbf{Model}} & \multicolumn{6}{c|}{\textbf{MIMIC}} & \multicolumn{6}{c}{\textbf{VinDr}} \\
\cmidrule(lr){2-7} \cmidrule(lr){8-13}
& \multicolumn{3}{c|}{\textbf{Yes/No}} & \multicolumn{3}{c|}{\textbf{Open-ended}} & \multicolumn{3}{c|}{\textbf{Yes/No}} & \multicolumn{3}{c}{\textbf{Open-ended}} \\ \cmidrule(lr){2-4} \cmidrule(lr){5-7} \cmidrule(lr){8-10} \cmidrule(lr){11-13}
& \textbf{F1} & \textbf{Precision} & \textbf{Recall} & \textbf{F1} & \textbf{Precision} & \textbf{Recall} & \textbf{F1} & \textbf{Precision} & \textbf{Recall} & \textbf{F1} & \textbf{Precision} & \textbf{Recall} \\
\midrule
\multicolumn{13}{c}{\cellcolor{lightgray!50} \textit{Proprietary Models}} \\ 
\midrule
Gemini-1.5-Flash-8B \cite{gemini} & 0.548 & 0.520 & 0.569 & 0.091 & 0.127 & 0.024 & 0.571 & 0.604 & 0.543 & 0.038 & 0.098 & 0.023 \\
GPT-4-O \cite{gpt4}               & 0.646 & 0.706 & 0.556 & 0.109 & 0.143 & 0.089 & 0.599 & 0.660 & 0.487 & 0.097 & 0.087 & 0.069 \\
GPT-4-Vision \cite{gpt4}          & 0.521 & 0.652 & 0.409 & 0.087 & 0.093 & 0.067 & 0.568 & 0.637 & 0.424 & 0.074 & 0.053 & 0.072 \\
\midrule
\multicolumn{13}{c}{\cellcolor{orange!20} \textit{Open-source Models}} \\
\midrule
CheXagent \cite{chen2024chexagent}       & 0.677 & 0.621 & 0.695 & 0.425 & 0.450 & 0.416 & 0.610 & 0.628 & 0.603 & 0.374 & 0.342 & 0.330 \\
BioMedGPT \cite{zhang2023biomedgpt}     & 0.708 & 0.734 & 0.672 & 0.397 & 0.387 & 0.412 & 0.624 & 0.606 & 0.631 & 0.289 & 0.265 & 0.310 \\
MedFlamingo \cite{alayrac2022flamingo}  & 0.562 & 0.574 & 0.552 & 0.365 & 0.390 & 0.310 & 0.584 & 0.600 & 0.547 & 0.301 & 0.285 & 0.322 \\
LLaVA-Med \cite{li2024llava}            & 0.722 & 0.712 & 0.709 & 0.551 & 0.541 & \underline{0.565} & 0.703 & 0.708 & 0.699 & 0.523 & 0.519 & 0.530 \\
LISA \cite{lai2024lisa}                 & 0.716 & 0.692 & 0.728 & 0.524 & 0.556 & 0.511 & 0.707 & 0.712 & 0.694 & 0.521 & 0.509 & \underline{0.534} \\
\midrule
\multicolumn{13}{c}{\cellcolor{yellow!20} \textit{Adversarial Techniques}} \\
\midrule
PH + LLaVA-Med \cite{zhang2024prompt}      & \underline{0.735} & \underline{0.754} & \underline{0.724} & \underline{0.563} & \underline{0.589} & 0.540 & 0.716 & 0.725 & 0.698 & 0.522 & 0.510 & 0.527 \\
VCD + LLaVA-Med \cite{leng2024mitigating}   & 0.716 & 0.708 & 0.721 & 0.566 & 0.583 & 0.538 & 0.652 & 0.699 & 0.591 & 0.501 & \textbf{0.547} & 0.497 \\
CRG + LLaVA-Med \cite{zhang2022contrastive} & 0.731 & 0.715 & 0.725 & 0.544 & 0.563 & 0.554 & 0.689 & 0.667 & 0.702 & 0.509 & 0.524 & 0.494 \\
\midrule
\rowcolor{green!20} LobA w/o self-prompt & 0.727 & 0.716 & 0.731 & 0.544 & 0.557 & 0.535 & \underline{0.724} & \underline{0.745} & 0.706 & 0.511 & 0.526 & 0.507 \\
\rowcolor{green!20} LobA (Ours)          & \textbf{0.752} & \textbf{0.759} & \textbf{0.743} & \textbf{0.581} & \textbf{0.593} & \textbf{0.572} & \textbf{0.728} & \textbf{0.758} & \textbf{0.711} & \textbf{0.542} & \underline{0.533} & \textbf{0.560} \\
\bottomrule
\end{tabular}%
}
\caption{Performance comparison across different models on the proposed HEAL-MedVQA benchmark on MIMIC and VinDr datasets for Yes/No and Open-ended VQA tasks. Text in \textbf{bold} and \underline{underline} highlights the best and second best results, respectively.}
\label{tab:performance_comparison}
\end{table*}}

\subsection{Attention Highlighter via Self-Prompting}
While the proposed training paradigm allows the model to localize the region of interest when answering, 
hallucination can still occur when the model pays little attention to the image compared to the text. To deal with this, we propose to highlight the attention on the queried regions by self-prompting the segmentation map. Our module consists of two functions: attention reweighting and contrastive decoding.

For \textit{attention reweighting}, patches of interest $x_{hl}$ are added extra weight before feeding through the softmax layer so that those regions get more focus during answer generation. Denote $h_{j,i}$ the original value of the attention logits of patch j to patch $i$, and $\beta$ the hyperparameter of added weight, then the new attention weight after reweighting $\hat{h}_{j,i}$ becomes
\begin{equation}
    \hat{h}_{j,i} =
\begin{cases} 
h_{j,i} & \text{if } i \notin x_{hl}, \\
h_{j,i} + \log \beta, \beta > 0 & \text{if } i \in x_{hl}.
\end{cases}
\end{equation}
Then the attention probability after Softmax is
\begin{equation}
    \tilde{a}_{j,i} = \frac{\beta^{m_i}\exp{(a_{j,i})}}{\sum_k{\beta^{m_k}\exp{a_{j,k}}}}.
\end{equation}
Here, $m_i=1$, indicating token $i$ is highlighted, and $m_i=0$ otherwise.
We apply this function to the visual backbone of M-LLM where the patches of interest are interpolated from the mask. Applying the reweighted attention $\tilde{A}$, we obtain the highlighted image tokens $\tilde{x}_{\text{img}}$:
\begin{equation}
    \tilde{x}_{\text{img}} = \tilde{A}V.
\end{equation}

To further alleviate shortcut learning, we perform \textit{contrastive decoding}~\cite{zhang2024prompt}, which contrasts the model decisions after and before highlighting regions of interest. Particularly, we compute the difference between the token probability after highlighting $p_{hl}$ and probability before highlighting $p_{bh}$ 
to obtain the decoded probability $p$:
\begin{equation}
\begin{split}
    p_{\text{bh}} & = p(\hat{y}_{\text{text}}|x_{\text{text}},x_{\text{img}})\\
    p_{\text{hl}} & = p(\hat{y}_{\text{text}}|x_{\text{text}},\tilde{x}_{\text{img}} )\\
    p & = \text{softmax}( (1+\alpha)\log p_{\text{hl}} - \alpha \log p_{\text{bh}} ).
\end{split}
\label{eq: cd}
\end{equation} where $\alpha$ is a hyperparameter controlling the degree of grounding. Applying contrastive decoding minimizes the effect of other tokens on the model output, preventing hallucination.

\noindent \textbf{Discussion.} 
Although Prompt Highlighter \cite{zhang2024prompt} also emphasizes the ``highlighting" of regions of interest to reduce hallucination, ours differs in two ways. First, Prompt Highlighter requires human prompting for attention highlighting, which is expensive in medical image analysis. In contrast, ours automatically localizes regions of interest, thus obviating the need for medical professionals to manually localize regions of interest. 
Second, while Prompt Highlighter only enhances attention on the language decoder, our methodology reweighs attention scores on the visual backbone. 
The deep intervention on the visual backbone is more effective in mitigating the visual shortcut learning. 


\section{Experimental Results}

In this section, we report the benchmark results of popular medical LMMs and our very own framework, as well as further analysis for our curated adversarial grounding VQA benchmark.

\begin{table}[hbt!] 
\scalebox{0.75}{
\centering
\begin{tabular}{lcccc}
\toprule
 & \multicolumn{2}{c}{MIMIC} & \multicolumn{2}{c}{VinDr} \\
\cmidrule(lr){2-3} \cmidrule(lr){4-5}
Method & Anatomy & Disease & Anatomy & Disease \\
\midrule
Gemini-1.5 & 0.532 & 0.510 & 0.521 & 0.487 \\
GPT-4-O & 0.594 & 0.603 & 0.509 & 0.596 \\
CheXagent & 0.621 & 0.650 & 0.689 & 0.650 \\
BioMedGPT & 0.687 & 0.610 & 0.623 & 0.612 \\
LLaVA-Med & 0.764 & 0.762 & 0.751 & 0.734 \\\midrule
PH + LLaVA-Med & 0.783 & 0.774 & 0.724 & \textbf{0.762}\\
VCD + LLaVA-Med & 0.744 & 0.788 & 0.740 & 0.750 \\
CRG + LLaVA-Med & 0.731 & 0.771 & 0.742 & 0.738 \\
\rowcolor{green!20} \textbf{LobA (Ours)} & \textbf{0.792} & \textbf{0.801} & \textbf{0.770} & 0.748 \\
\bottomrule
\end{tabular} }
\caption{Textual Perturbation Test on MIMIC and VinDr datasets} 
\label{tab:lang_bias}
\end{table}
\begin{table}[htb]
\centering
\scalebox{0.8}{ 
\begin{tabular}{lcccc}
\toprule
 & \multicolumn{1}{c}{MIMIC} & \multicolumn{1}{c}{VinDr} \\
\midrule
Gemini-1.5 & 0.410 & 0.403 \\
GPT-4-O & 0.557 & 0.581  \\
CheXagent & 0.631 & 0.671 \\
BioMedGPT & 0.647 & 0.653 \\
LLaVA-Med & 0.696 & 0.680  \\\midrule
PH + LLaVA-Med & 0.720 & 0.698 \\
VCD + LLaVA-Med & 0.678 & 0.654  \\
CRG + LLaVA-Med & 0.682 & 0.693  \\
\rowcolor{green!20}\textbf{LobA (Ours)} & \textbf{0.734} & \textbf{0.701}  \\
\bottomrule
\end{tabular} }
\caption{Visual Perturbation Test on MIMIC and VinDr datasets}
\label{tab:visual_bias}
\end{table}

\begin{table}[h!]
\centering
\scalebox{0.75}{
\begin{tabular}{llccc}
\toprule
\multirow{2}{*}{\textbf{Module}} & \multirow{2}{*}{} & \cellcolor{gray!25}\textbf{Backbone} & \cellcolor{green!25}\textbf{LobA w/o} & \cellcolor{green!25}\textbf{LobA} \\
& &  \cellcolor{gray!25} & \cellcolor{green!25}\textbf{self-prompt} & \cellcolor{green!25}\textbf{(Ours)} \\
\midrule
\multirow{2}{*}{Component} 
& Localization & \cellcolor{gray!25}{\textcolor{red}{\XSolidBrush}} & \cellcolor{green!25}\textcolor{Green}{\CheckmarkBold} & \cellcolor{green!25}\textcolor{Green}{\CheckmarkBold} \\
& Self-prompt & \cellcolor{gray!25}{\textcolor{red}{\XSolidBrush}} & \cellcolor{green!25}{\textcolor{red}{\XSolidBrush}} & \cellcolor{green!25}\textcolor{Green}{\CheckmarkBold} \\
\midrule
\multirow{2}{*}{Perb Test} 
& TPT & \cellcolor{gray!25}0.763 & \cellcolor{green!25}0.772 & \cellcolor{green!25}\textbf{0.797} \\
& VPT & \cellcolor{gray!25}0.743 & \cellcolor{green!25}0.738 & \cellcolor{green!25}\textbf{0.759} \\
\midrule
\multirow{4}{*}{F1 Score} 
& MIMIC (Yes/No) & \cellcolor{gray!25}0.722 & \cellcolor{green!25}0.727 & \cellcolor{green!25}\textbf{0.752} \\
& MIMIC (Open-ended) & \cellcolor{gray!25}0.551 & \cellcolor{green!25}0.544 & \cellcolor{green!25}\textbf{0.581} \\
& VinDr (Yes/No) & \cellcolor{gray!25}0.703 & \cellcolor{green!25}0.724 & \cellcolor{green!25}\textbf{0.728} \\
& VinDr (Open-ended) & \cellcolor{gray!25}0.523 & \cellcolor{green!25}0.511 & \cellcolor{green!25}\textbf{0.542} \\
\bottomrule
\end{tabular} }

\caption{Ablation studies of different components of LobA. The TPT and VPT scores are averages for MIMIC and VinDr datasets.}
\label{tab:ablation_studies}
\end{table}

\noindent \textbf{Benchmarks.} We conduct a thorough and systematic evaluation of the most popular Medical Large Multimodal Models on our large-scale HEAL-MedVQA benchmark. Specifically, we utilize 7 state-of-the-art multi-modal LLMs for our evaluation protocol, 3 of which are proprietary models: GPT-4o, GPT-4 Vision \cite{gpt4} and Gemini 1.5 \cite{gemini} and 4 are open source M-LLMs of the medical domain: CheXagent \cite{chen2024chexagent}, LLaVA-Med \cite{li2024llava}, BioMedGPT \cite{zhang2023biomedgpt} and Med-Flamingo \cite{alayrac2022flamingo}. We also assess the capabilities of state-of-the-art adversarial VQA methods, including VCD \cite{leng2024mitigating}, Prompt Highlighter \cite{zhang2024prompt}, and CRG \cite{crg}. 

\noindent \textbf{Implementation Details.}
Open source multi-modal LLMs are fine-tuned on our training dataset with LoRA \cite{hu2021lora} before evaluation. The learning rate is tuned from the range of values $\{1e-4, 1e-5, 1e-6\}$. For LobA framework, we used LLaVA-Med \cite{li2024llava} as the pretrained MLLM and MedSAM \cite{kirillov2023segment} as the segmentation model. Following \cite{zhang2024prompt}, we select $\alpha = 0.3$, $\beta_{ViT} = \beta_{LLM} = 2$ as the set of hyperparameters for the Attention Highlighter module.
All models are fine-tuned using Transformer, PyTorch and DeepSpeed frameworks on a single 80GB A100 GPU cluster.  
During inference, the temperature of all multi-modal LLMs is set to 0.1, and the beam size is set to 1.\\
\noindent \textbf{Evaluation Metrics.}
To assess the accuracy of the multi-modal LLMs, we utilize  LLaMa 3.1-8B to extract information from the M-LLM's output, and categorize them into a set of disease labels. For example, for an open-ended question, if the model returns the answer \textit{``The heart suffers from pneumonia, pulmonary fibrosis and nodule/mass"} then the extracted labels by the LLM will be \textit{pneumonia}, \textit{pulmonary fibrosis}, \textit{nodule/mass}. We report the multi-label precision, recall, and micro F1 score as the main accuracy metrics. For hallucination sensitivity analysis, we adopt our proposed evaluation metrics, Textual Perturbation Test (TPT), and Visual Perturbation Test (VPT) scores, as discussed in Sec.~\ref{sec:protocol}.

\noindent \textbf{Experimental Results and Discussion.}
\textit{VQA Accuracy.}
As shown in Table \ref{tab:performance_comparison}, proprietary models achieve sub-optimal accuracy. Most models have accuracy less than 50\% on binary Yes/No questions and less than 10\% 
for open-ended questions. This shows the challenging nature of our HEAL-MedVQA benchmark, requiring the model to be robust against visual and textual shortcut learning. 
Our proposed framework consistently outperforms recent advances in M-LLMs and adversarial VQA techniques even after fine-tuning them.
Notably, the proposed LoBA outperforms the state-of-the-art Prompt Highlighter \cite{zhang2024prompt} by 3.19\% and 3.63\% 
on Open-ended questions for VinDr and MIMIC-CXR, respectively. Injecting grounding prior, as in our proposal, boosts the VQA accuracy. 

 \textit{Textual Perturbation Test.} Table \ref{tab:lang_bias} reports the sensitivity of each model to textual language bias. 
 The higher the percentage is, the more robust a model is against textual shortcut bias.
Our proposed framework achieves the highest sensitivity measure overall, with the ratio of changed answers drastically increasing by up to 3.89\% 
compared to CRG \cite{crg}.
Enforcing visual reasoning via highlighted attention to the queried regions, as in our proposal, alleviates the shortcut learning.
Lastly, fine-tuned open-source models are generally more robust than closed-source models like GPT-4-O, showing that fine-tuning LLMs on our adversarial data remedies hallucination. 

\textit{Visual Perturbation Test.} Table \ref{tab:visual_bias} reports the visual sensitivity score, with the score defined as the number of changed answers when the original image is overlaid with a healthy localized anatomy.
Most multi-modal LLMs in our benchmark shift their answer around 40-70\% of the time. 
Our method is the most sensitive to the change in the localized area, increasing the number of visually dependent answers by 1.94\% compared to Prompt Highlighter~\cite{zhang2024prompt}.

\textit{Ablation studies.} Table \ref{tab:ablation_studies} reports the effects of each component in our framework: the LLaVA-Med backbone, Grounded Text Generation (GTG) module, and the full framework with self-prompting. Applying GTG improves the model's performance in some categories, showing the effectiveness of localization-aware training. 
With the self-prompting framework, the accuracy and textual/visual robustness improve greatly, most evidently increasing 4.45\%
in TPT score, and 2.73\% on MIMIC-CXR dataset Yes/No F1 Score. 

\begin{figure}
    \centering
    \includegraphics[width=1.0\linewidth]{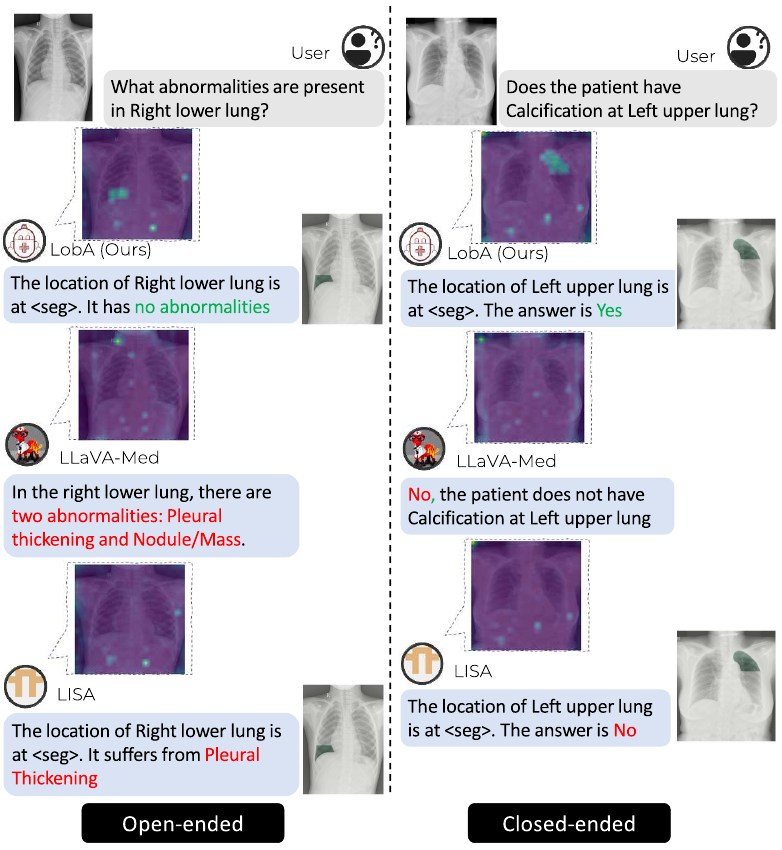}
    \caption{Qualitative case study of both question types}
    \label{fig:qualitative}
\end{figure}

\textit{Qualitative analysis.}
Fig. \ref{fig:qualitative} showcases our \textit{LobA} and some other LMM's response on some qualitative examples, as well as the visual attention maps on the input images. We can see that \textit{LobA} manages to query the attention weights much more intuitively compared to LLaVA-Med or LISA. While others' attentions are scattered into many irrelevant areas, in the first example LobA's main attention weights are allocated around the right lower lung area, and the second example is near the left upper lung. Thanks to the grounding generation module, followed by LobA's self-prompting module to automatically re-weight the attention to the area of interest, the model is able to have better visual reasoning capabilities, 
leading to more accurate answers compared with other methods.

\section{Conclusion}

This paper introduces Heal-MedVQA, a new large-scale Medical Visual Question Answering Benchmark with over 67,000 question-answer pairs, which queries diseases at local anatomies, evaluates LMM's capabilities to localize at grounded visual evidence when answering. In addition, we present the Localize-before-Answer (LobA) framework, which trains the LMMs to segment the region of interest and re-adjust their attention for more emphasis on the segmented pathological areas, leading to more reliable answers. Our experimental results showed that our framework LobA outperformed state-of-the-art medical LMMs on Heal-MedVQA, proving its robustness and localization capabilities.


\newpage
\appendix
\section{Computational Efficiency}
We present the inference time (seconds) and GPU consumption (MB) of \textit{LoBA} on a system with a single A100 80GB GPU for 100 samples in comparison with some other medical LMMs in Tab.~\ref{tab:compare}. Our LobA yields computational efficiency comparable to that of existing methods, while achieving superior localization capabilities.

\begin{table}[h!]

\centering
\small 
\resizebox{1.0\linewidth}{!}{ 
\begin{tabular}{lcccc}
\hline
& CheXagent & Med-Flamingo & LLaVA-Med & \cellcolor{highlightgreen}LobA (Ours) \\
\hline
Inf. Time & 113.4 & 126.5 & 87.0 & \cellcolor{highlightgreen}101.2 \\
GPU Mem   & 23406 & 26453 & 17729 & \cellcolor{highlightgreen}25254 \\
\hline
\end{tabular}
}
\vspace{-1mm}
\caption{Comparison of inference time and GPU consumption}
\label{tab:compare}
\end{table}

\section{Benchmarking Other General LLMs}
We report the F1 score of additional proprietary and open-source LMMs (fine-tuned) on the VinDr subset in Tab.  \ref{tab:general_models}

\begin{table}[h!]
\centering
\small
\resizebox{1.0\linewidth}{!}{%
\begin{tabular}{lccccc}
\hline
\textbf{Model} & Gemini-1.5 & Claude 3.7 & Qwen 2.5 VL & InternVL & \cellcolor{highlightgreen}\textbf{LobA} \\
& Pro & Sonnet & 7B-Instruct & 2.5 8B & \cellcolor{highlightgreen}\textbf{(Ours)}\\
\hline
Yes/No & 0.583 & 0.610 & 0.660 & 0.654 & \cellcolor{highlightgreen}\textbf{0.728} \\
Open   & 0.078 & 0.062 & 0.413 & 0.378 & \cellcolor{highlightgreen}\textbf{0.542} \\
\hline
\end{tabular}%
}

\caption{Performance on VinDr for general LMMs}
\label{tab:general_models}
\end{table}

\section{Ablation study}
\begin{figure}[h!]
\centering
\begin{minipage}{0.17\textwidth}
    \centering
    \small
    \renewcommand{\arraystretch}{1.2}
    \resizebox{\linewidth}{!}{
    \begin{tabular}{l|c|c}
    \toprule
    $\beta$ & Yes/No F1 & Open F1 \\
    \midrule
    0.5  & 0.694 & 0.476\\
    1  & 0.698 & 0.510\\
    2 & 0.720 & 0.545 \\
    3  & 0.728 & 0.542\\
    4  & 0.712 & 0.501\\
    \bottomrule
    \end{tabular}
    }
    \vspace{5mm}

    \captionof{table}{Effect of $\beta$ on performance on VinDr}
    \label{tab:alpha}
\end{minipage}%
\hspace{3mm}
\begin{minipage}{0.28\textwidth}
    \centering
    \includegraphics[width=0.75\linewidth]{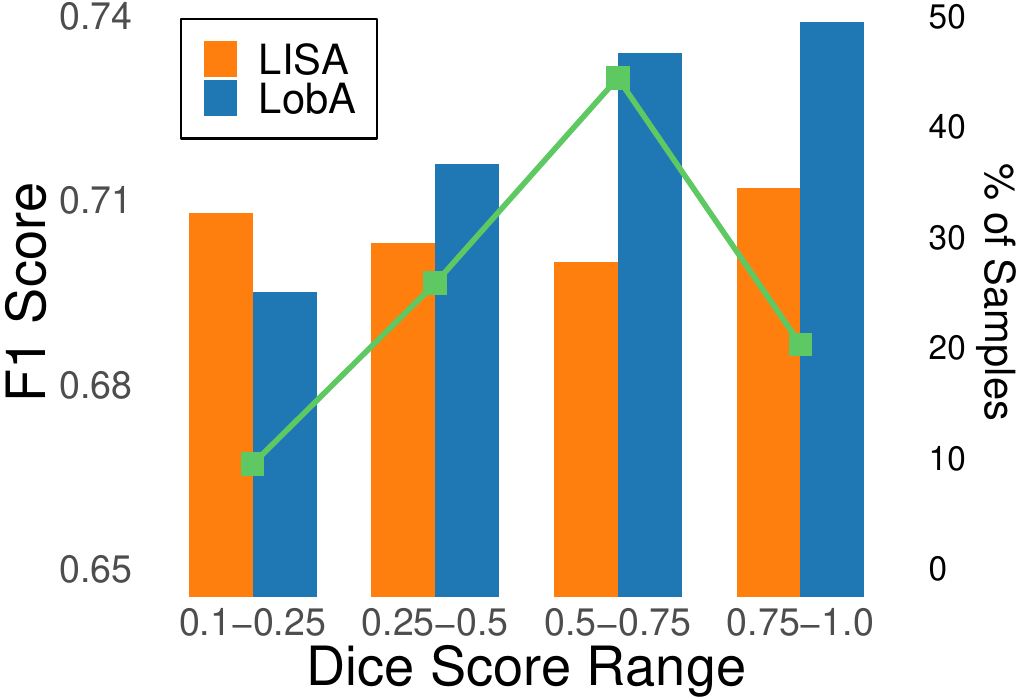}
    \captionof{figure}{Yes/No F1 w.r.t segmentation qualities. Data portion (\%) on test set is also shown.}
    \label{fig:segmentation_ablation}
\end{minipage}

\end{figure}

We conducted two ablation studies, examining the effect of added attention weights $\beta$ (Tab. \ref{tab:alpha}) and the effect of segmentation quality on the model's results (Fig \ref{fig:segmentation_ablation}).\\ 
In summary, low weights cause the segmented area to have minimal influence, harming results, while excessive weight (e.g., 4.0) shifts focus away from the question, also degrading performance. For the second ablation study, the quality of the segmentation indeed correlates directly with \textit{LoBA}'s performance. Overall, our model outperforms LISA baselines when the segmentation Dice score $> 0.25$, which accounts for more than $10\%$ of the test set. This shows that our design is already beneficial even when the segmentation module can localize the rough regions to attend to. For samples with Dice $< 0.25$, which only account for 9.47\% of data, our LoBA yields a slight performance decrease.

\section{LobA Prompt Template}

\begin{figure}[!htb]
    \centering
    \includegraphics[width=1\linewidth]{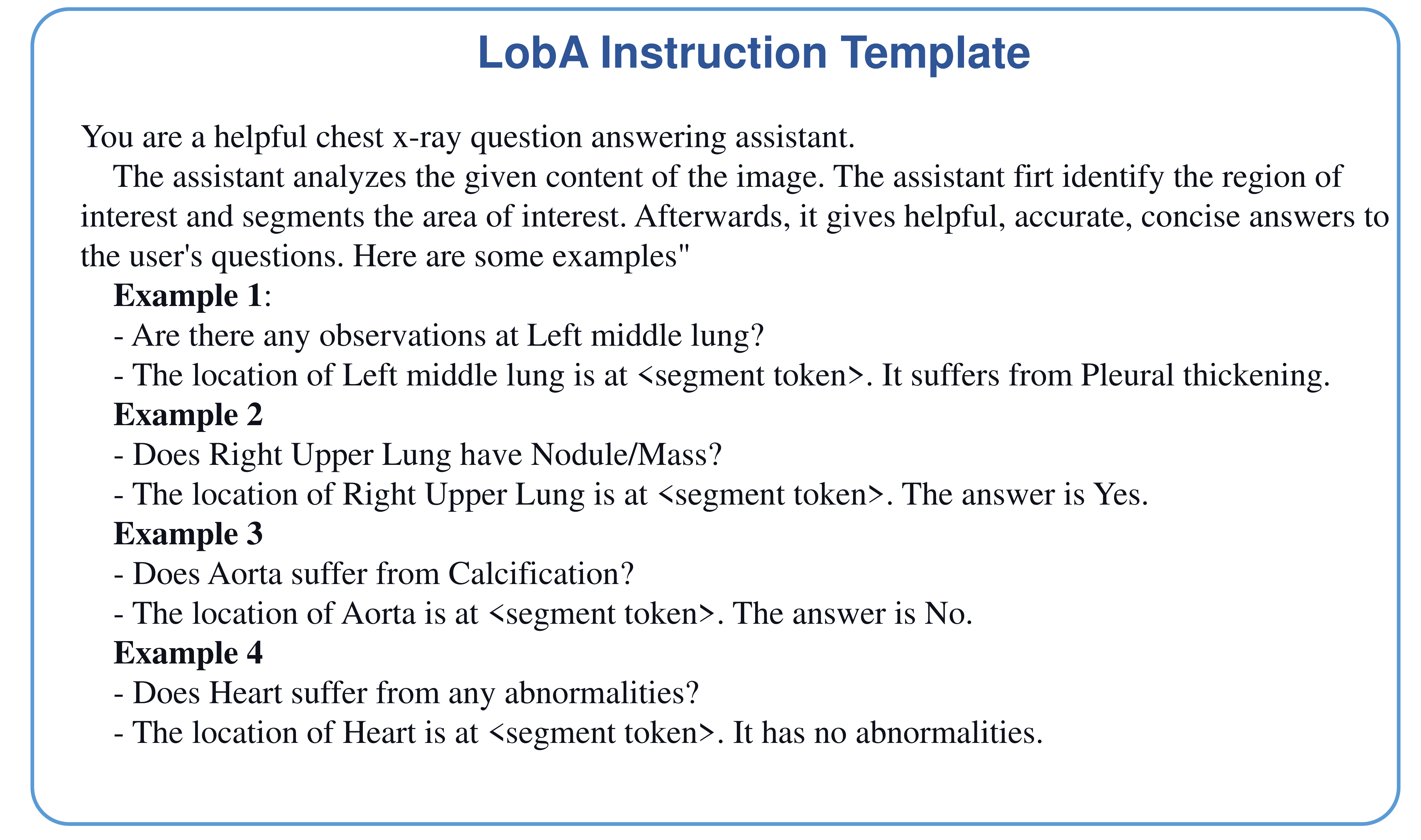}
    \caption{LobA Instruction Template}
    \label{fig:loba_template}
\end{figure}

\begin{figure}[!htb]
    \centering
    \includegraphics[width=1\linewidth]{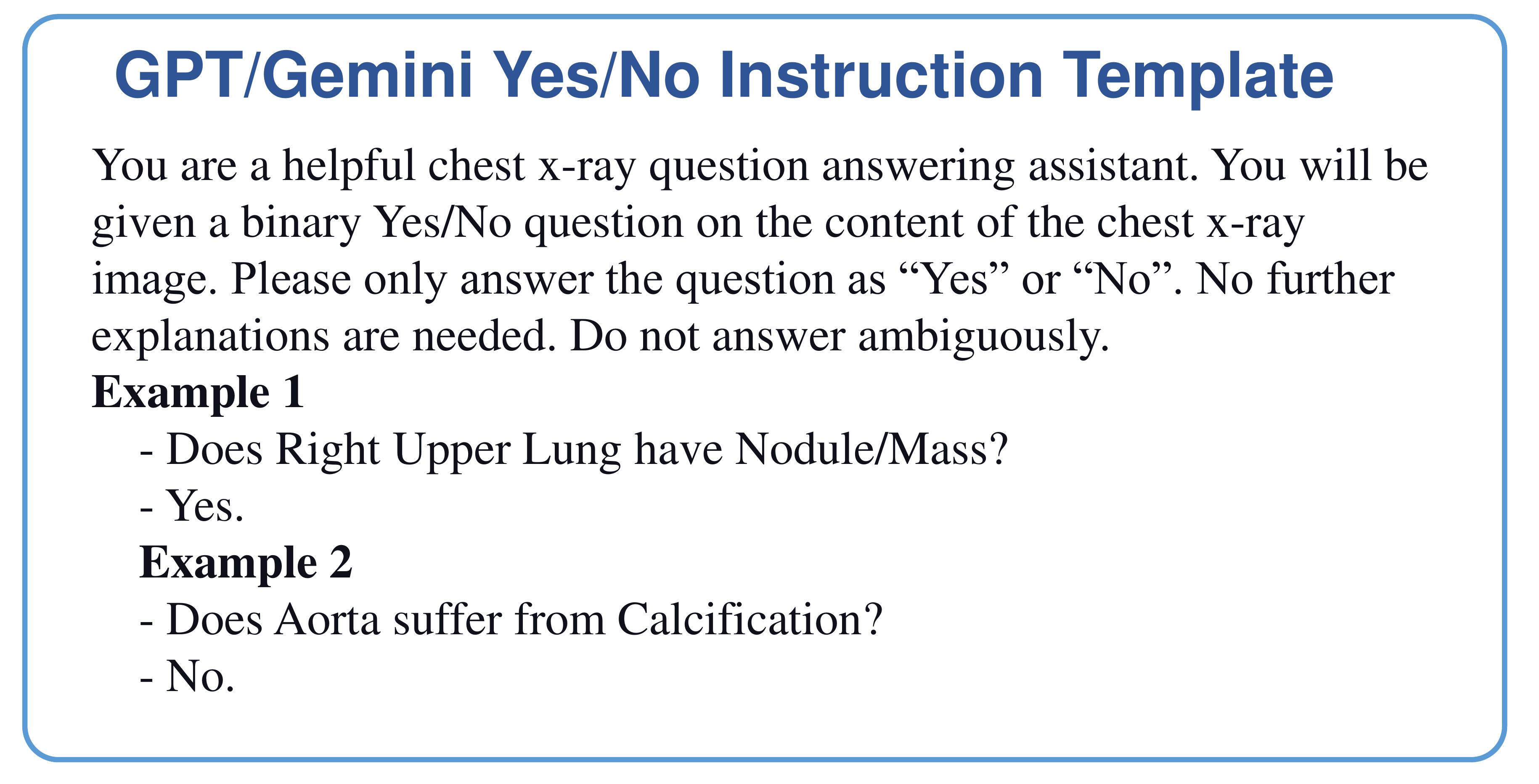}
    \caption{GPT and Gemini Models Instruction Template for Yes/No questions}
    \label{fig:gpt_template_yesno}
\end{figure}

\begin{figure}[!htb]
    \centering
    \includegraphics[width=1\linewidth]{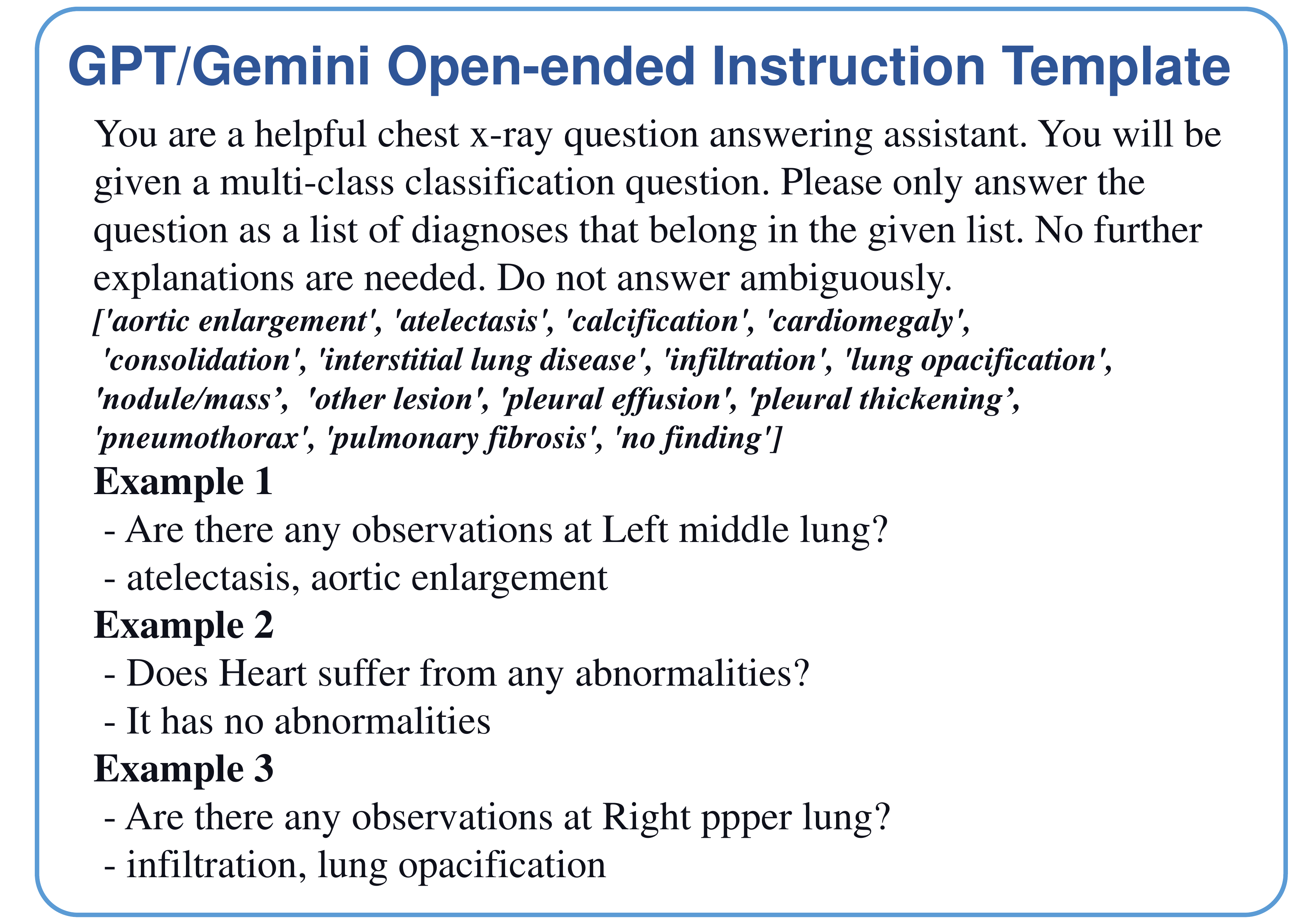}
    \caption{GPT and Gemini Models Instruction Template for Open-ended questions}
    \label{fig:gpt_template_open}
\end{figure}

\begin{figure*}
    \centering
    \includegraphics[width=1\linewidth]{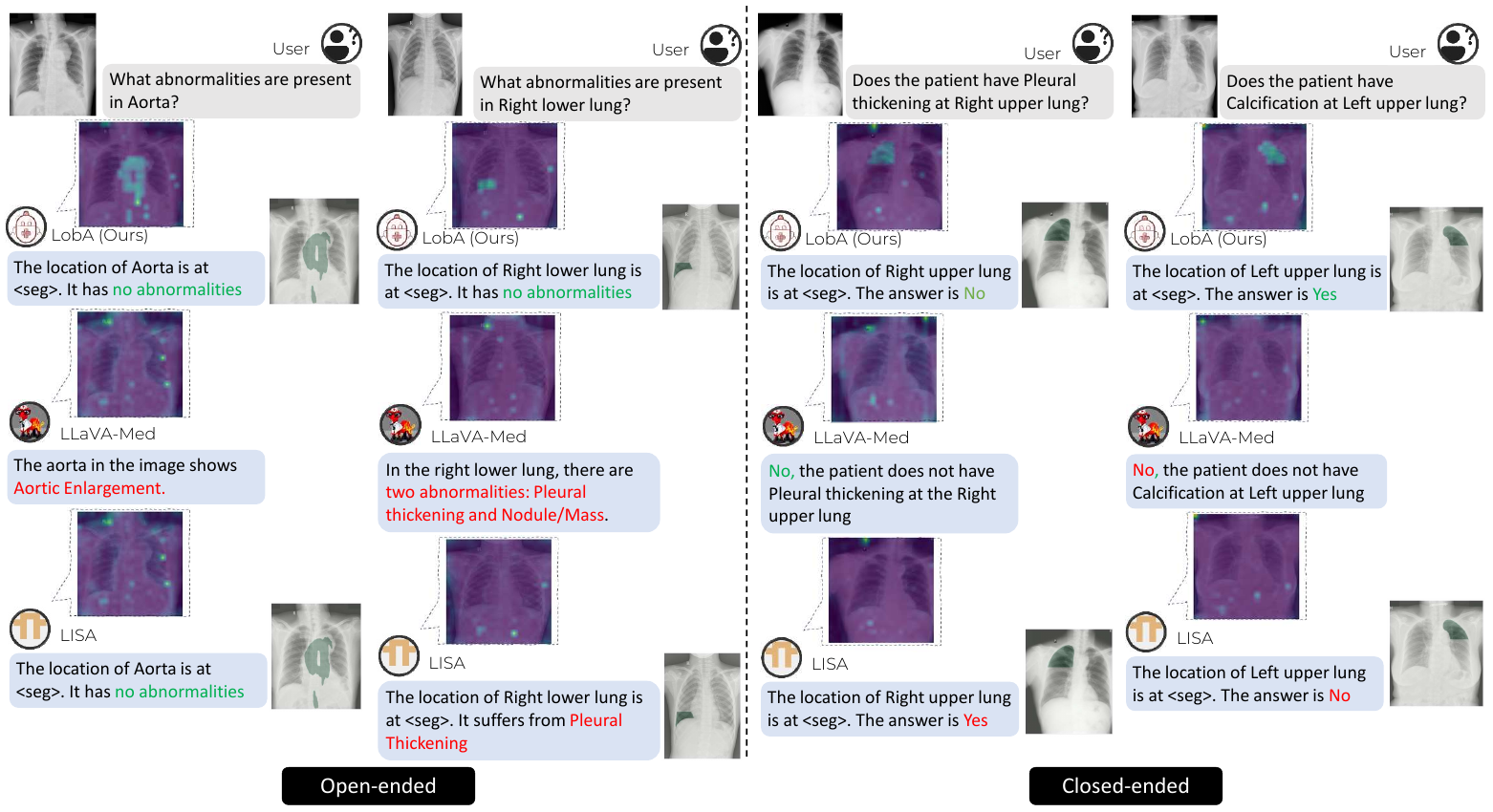}
    \caption{Qualitative Examples from our dataset}
    \label{fig:qualitative2}

\end{figure*}

We propose a Grounded Text Generation training paradigm to enable LMM to localize the area of interest with the segmentation token ``[SEG]``, before giving its answer for the model to be aware of the localization information. \\
As such, our model is fine-tuned to follow a strict template as 
\textit{``The location of \{relevant areas\} is at \textless SEG\textgreater. The answer is ..."}. To make sure the model's output comply with this standard, we designed a few-shot instruction template for LobA so that the model can follow the desired template more accurately, displayed in Fig \ref{fig:loba_template}

\section{Proprietary Models Template}
For GPT-4-O and GPT-4-Vision and Gemini-1.5-Flash-8B  models, we modify the template of these proprietary models for yes/no questions and open-ended questions separately to allow them to follow our instructions correctly. These 2 templates are displayed by Fig \ref{fig:gpt_template_open} and \ref{fig:gpt_template_yesno}

\section{Evaluation Template}

\begin{figure}[!htb]
    \centering
    \includegraphics[width=1\linewidth]{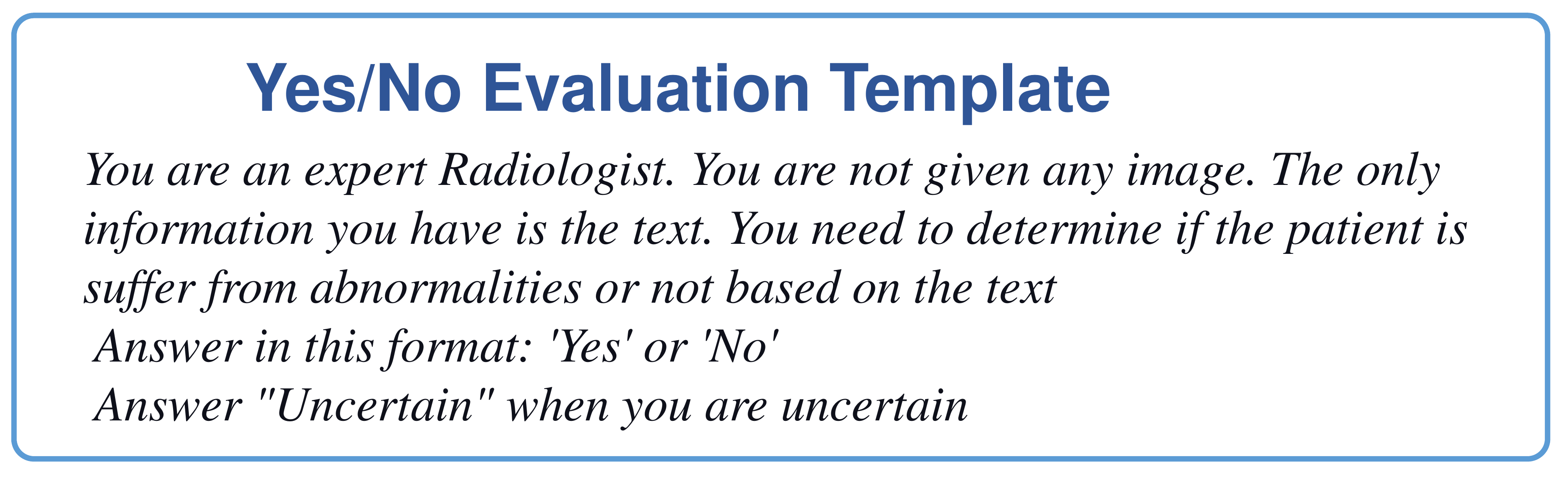}
    \caption{Yes/No evaluation prompt template}

    \label{fig:yesno_eval}
\end{figure}

\begin{figure}[!htb]
    \centering
    \includegraphics[width=1\linewidth]{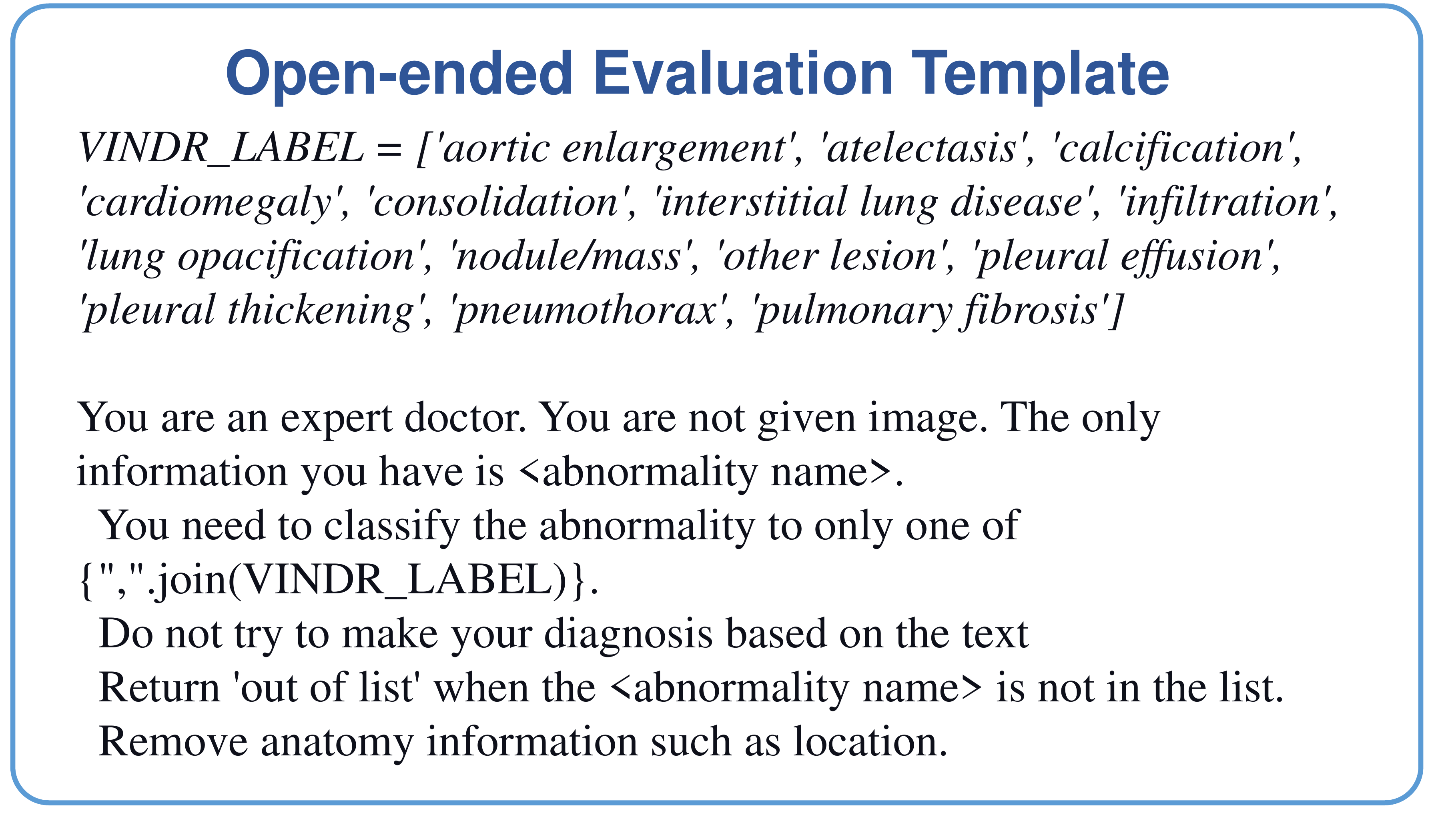}
    \caption{Open Ended evaluation prompt template}

    \label{fig:open_ended_eval}
\end{figure}

We utilize LLaMA-3 as the LLM backbone to extract the labels from the model's natural language answers with the following prompt templates, Fig \ref{fig:yesno_eval} and Fig \ref{fig:open_ended_eval}


\section{Question example and qualitative results}
We present some qualitative examples, as well as some model's response and their visual attention map in Fig \ref{fig:qualitative2}.

\section*{Acknowledgements}
We would like to thank Jiangyu Zhou for assisting with the annotation of chest X-ray images for this project.
\bibliographystyle{main}
\bibliography{main}

\end{document}